\documentclass{article}
\usepackage{amsmath,epsfig}
\usepackage[preprint]{spconfa4}
\usepackage{booktabs}
\usepackage{amsfonts}
\usepackage{fancyhdr}
\usepackage[pagebackref=true,breaklinks=true,letterpaper=true,colorlinks,bookmarks=false]{hyperref}

\copyrightnotice{978-1-6654-3864-3/21/\$31.00 2021 IEEE}

\let\OLDthebibliography\thebibliography
\renewcommand\thebibliography[1]{
  \OLDthebibliography{#1}
  \setlength{\parskip}{0pt}
  \setlength{\itemsep}{0pt plus 0.3ex}
}

\begin{document}\sloppy

\def\x{{\mathbf x}}
\def\L{{\cal L}}

\title{SDAN: Squared Deformable Alignment Network for Learning Misaligned Optical Zoom}
%
\name{Kangfu Mei, Shenglong Ye, Rui Huang$^{\ast}$\thanks{$^{\ast}$Corresponding author. This work is supported in part by funding from Shenzhen Institute of Artificial Intelligence and Robotics for Society, and Shenzhen NSF JCYJ20190813170601651.}}
\address{\{kangfumei, shenglongye\}@link.cuhk.edu.cn, ruihuang@cuhk.edu.cn \\Shenzhen Institute of Artificial Intelligence and Robotics for Society, \\The Chinese University of Hong Kong, Shenzhen}

\maketitle
\begin{abstract}
  Deep Neural Network (DNN) based super-resolution algorithms have greatly improved the quality of the generated images.
  However, these algorithms often yield significant artifacts when dealing with real-world super-resolution problems due to the difficulty in learning misaligned optical zoom.
  In this paper, we introduce a Squared Deformable Alignment Network (SDAN) to address this issue.
  Our network learns squared per-point offsets for convolutional kernels, and then aligns features in corrected convolutional windows based on the offsets.
  So the misalignment will be minimized by the extracted aligned features.
  Different from the per-point offsets used in the vanilla Deformable Convolutional Network (DCN), our proposed squared offsets not only accelerate the offset learning but also improve the generation quality with fewer parameters.
  Besides, we further propose an efficient cross packing attention layer to boost the accuracy of the learned offsets.
  It leverages the packing and unpacking operations to enlarge the receptive field of the offset learning and to enhance the ability of extracting the spatial connection between the low-resolution images and the referenced images.
  Comprehensive experiments show the superiority of our method over other state-of-the-art methods in both computational efficiency and realistic details. Code is available at \href{https://github.com/MKFMIKU/SDAN}{https://github.com/MKFMIKU/SDAN}
\end{abstract}
\section{Introduction}
\label{sec:intro}

DNN based image super-resolution methods~\cite{dong2015image, johnson2016perceptual, ledig2017photo, lim2017enhanced} have achieved significant progress on simulated super-resolution datasets, which consist of low and high resolution image pairs simulated by interpolation.
These methods learn to parameterize mappings between low resolution images and high resolution images, and they are optimized by using the measured distance between generated images and corresponding ground truth as the loss function.
However, limited by the learning on the simulated images, which cannot fully model the sophistication of the real-world optical zoom, these methods tend to add unrealistic details on the real-world images.
Hence, more recent works~\cite{zhang2019zoom,cai2019toward} propose to learn the super-resolution mappings on low and high resolution image pairs collected under nature environments.
Specifically, such image pairs are collected using a zoom lens.
The lens is moved into different places with different lens settings to capture images for the same major objects.
Due to the different distances between the lens and the major objects, the captured major objects in different images own different resolutions, which represent different levels of optical magnification under real-world conditions. We call such datasets optical zoom datasets.
Experiments in the previous work~\cite{zhang2019zoom} have proven that models learned on optical zoom datasets can generate images with higher visual quality, compared to the methods learned on simulated datasets.

However, even though image pairs taken under different focal lengths represent more accurate real-world mappings, they may lead to another crucial issue, i.e., the misalignment between the high-resolution and low-resolution images.
The misalignment comes from the viewpoint movements from the major objects, as well as the optical distortion in the lens.
Such unavoidable factors cause pixel shift or perspective change, hence the major objects in the low-resolution images have different magnified coordinates from the same objects in the corresponding high-resolution images.
When directly learning on the misaligned optical zoom datasets with pixel-wise loss functions, the algorithm will calculate incorrect distances and hence error gradients for parameter optimization.
According to the definition of the widely used pixel-wise loss function $\mathcal{L}_1$, the distance between the generated image $\tilde{X}$ and its ground truth $Y$ is formulated as:
\begin{equation}
  \mathcal{L}_1 (\tilde{X}, Y) = \frac{1}{N}\sum_i^N|\tilde{X}_i - Y_i|,
\end{equation}
where $N$ is the number of pixels in the generated image $\tilde{X}$.
When the pixels are misaligned, e.g., assuming the misalignment is a coordinate mapping function $F_R(*)$, the actual pixel-wise loss $'\mathcal{L}_1$ should be:
\begin{equation}
  '\mathcal{L}_1 (\tilde{X}, Y) = \frac{1}{N}\sum_i^N|\tilde{X}_{F_R(i)} - Y_i|,
\end{equation}
yet by using $\mathcal{L}_1$ instead of $'\mathcal{L}_1$, incorrect distance measurement will mislead the optimization of the DNN parameters.
The generated results from such networks will usually be over-smoothed as Figure~\ref{fig:img} shows.

To eliminate the possible problems of the pixel-wise loss functions, the recently proposed Contextual~\cite{mechrez2018contextual} loss function and its extended Contextual Bilateral~\cite{zhang2019zoom} loss function calculate the distance on the feature level instead of the pixel level. To be specific, the vanilla Contextual loss function is formulated as:
\begin{equation}
  \mathcal{L}_{CX} (\tilde{X}, Y) = \frac{1}{N}\sum_i^N \min_{j=1,\dots,M} (\Psi(\tilde{X})_i, \Psi(Y)_j),
\end{equation}
where $\Psi(*)$ is the pre-trained network for feature extraction, usually VGG19, and $N, M$ are the numbers of points in the extracted feature maps respectively.
Though the distance measurement in such a form can avoid the pixel-wise misalignment issue, we find that directly calculating distances between the extracted features leads to color offset, i.e., the generated images have incorrect colors in both the entire scenes and the specific objects, as Figure~\ref{fig:img} shows.
The calculation in the contextual space also costs more computational resources during training.
Therefore, compared with the pixel-wise loss functions, the feature-level contextual loss functions are still not as satisfactory as we would want them to be.

Inspired by Tian et al.~\cite{tian2020tdan} and Chan et al.~\cite{chan2020understanding}, which applied Deformable Convolutional Networks (DCNs)~\cite{dai2017deformable, zhu2019deformable} to neighboring frame alignment, in this paper, we propose to utilize DCN for learning optical zoom on misaligned images.
However, due to the differences between the two tasks, we find that directly applying DCN to misaligned optical zoom cannot achieve desired gains.
To elaborate, neighboring frame alignment takes the misaligned neighboring frames as well as the source frame as input, and it learns to align the neighboring frames especially the local contents within the frames, which requires subtle offset estimation during alignment.
In contrast, the misalignment in the optical zoom learning problem is caused by the lens movement or optical aberrations, while the contents are usually fixed in the scenes. Therefore we argue that our problem requires offset estimation on a larger yet coarser scale.


In this paper, we propose a new Squared Deformable Alignment Network (SDAN) to address these issues in the misaligned optical zoom task.
We introduce the squared offset that represents the offset of the whole convolutional window instead of the offset of each point in the window.
It is a more efficient strategy for feature alignment and greatly reduces the difficulties in learning optical zoom.
Furthermore, it also yields clearer results during inference because the learned offsets have lower diversity, which reduces the possibility of overfitting in the later convolutional layers.
To increase the limited receptive field of the vanilla DCN, which is not suitable for estimating global offsets, we introduce an efficient attention block, called Cross Packing Attention (CPA). It first applies a space-to-depth operation to pack the features into small pieces, and then uses channel attention on the packed features to extract cross-channel and spatial connections in an efficient way. 
Since the packing operation on images transfers the squared neighboring features into the same channel, the channel attention after the operation is able to extract feature relationships in the spatial neighboring area effectively.
The proposed CPA is more suitable for optical zoom learning than other attention mechanisms because the corresponding pixels in misaligned images usually can be found in the neighboring areas instead of the entire images~\cite{mechrez2018contextual}.
By combining the above two contributions, the proposed SDAN is able to efficiently learn optical zoom on the misaligned optical zoom datasets using pixel-wise loss functions.

\section{Related Work}
\subsection{DNN based Image Super-Resolution}
Based on the simulated low and high resolution images pairs, Dong et al.~\cite{dong2015image} first proposed SRCNN that represented mappings between low and high resolution images as a deep convolutional neural network.
The network is optimized by minimizing the average difference in corresponding pixels between the output and the target.
Following such a manner, SRCNN as well as its variants, e.g., SRResNet~\cite{ledig2017photo}, EDSR~\cite{lim2017enhanced}, MSRN~\cite{li2018multi}, gained great improvement on the simulated super-resolution benchmarks.
However, these methods tend to generate high-resolution images with unrealistic details.
To address this issue, Zhang et al.~\cite{zhang2019zoom} and Cai et al.~\cite{cai2019toward} proposed to train deep models on real-world low and high resolution image pairs collected with different focal lengths.
Networks trained on such datasets have better robustness than those trained on simulated datasets for real-world low-resolution images.
Yet these datasets are collected under human operations, which are not precise enough to avoid lens movement. There are also unavoidable optical aberrations. Both lead to misalignment between low and high resolution images.
Zhang et al.~\cite{zhang2019zoom} utilized contextual loss~\cite{mechrez2018contextual} and modified contextual bilateral loss to search the minimal difference between each point on the generated images and its label on the fly, which is able to avoid the misalignment issue but is also limited by the huge computation, so the images and labels are compacted into small dimensions with a pre-trained VGG network.
However, we observed that the implicit difference measurement between features instead of pixels might lead to color offset.

\begin{figure*}[htbp]
  \centering
  \includegraphics[scale=0.6]{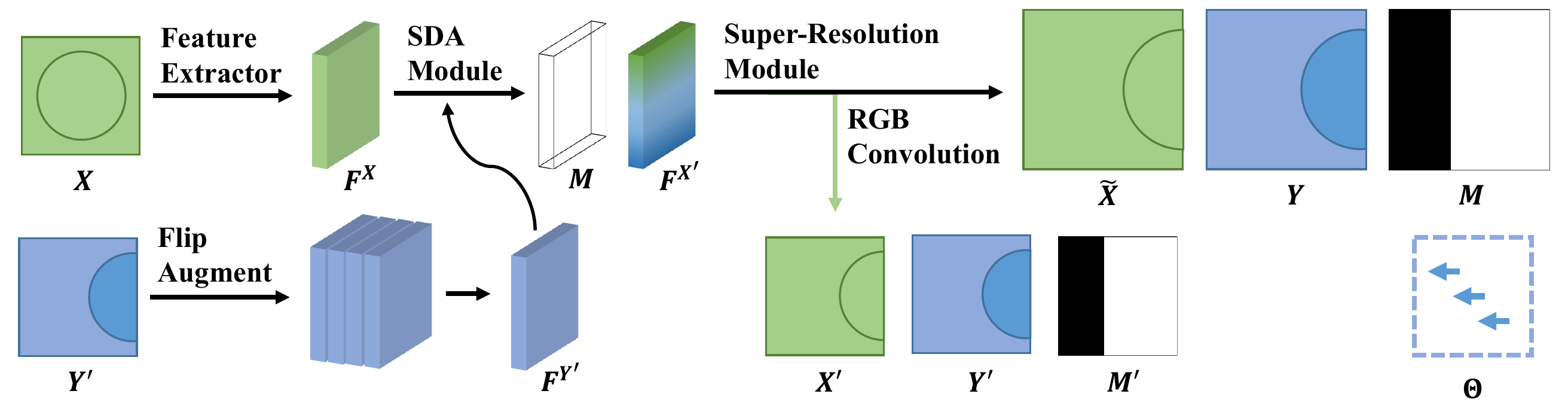}
  \caption{Overview of our proposed SDAN. Two graphs that contain a circle in different places are used to denote a misaligned low-high resolution image pair. SDAN takes low-resolution image $X$ and reference image $Y'$ as input then generates the temporally aligned result $X'$, final zoomed result $\tilde{X}$, and mask map $M$. The learned offset $\Theta$ is illustrated to denote offsets.}
  \label{fig:pipeline}
\end{figure*}

\subsection{Image/Feature Alignment}
Another solution to the misalignment issue is to align the images or features before calculating differences.
Some recent works~\cite{cai2019toward, mei2019higher} are trained on images that captured static scenes or printed scenes on papers to avoid large movement, and they used a handcrafted iterative registration algorithm or a sliding window based strategy to align the images.
However, these registration/alignment algorithms can only ensure the center patch of the images to be aligned yet other parts are ignored, which leads to high cost in data collection.
Learning to align features is a type of more feasible methods, especially the multi-frame based methods that learn to align neighboring frames to establish inter-frame correspondence.
~\cite{yi2019progressive, sajjadi2018frame, tao2017detail, isobe2020videoa, isobe2020videob} apply the learned optical flow to warp the neighboring frames, then pass the warped frames to deep networks for feature combining.
More recent works ~\cite{wang2018esrgan, tian2020tdan, chan2020understanding} found that learning the offsets of convolution kernels instead of pixels led to better performance and more natural details, in which the learned offsets control the convolutional windows of the regular convolutional networks in DCN~\cite{dai2017deformable}.
However, the above methods are proposed for combining neighboring frames, which are not robust enough to be applied to optical zoom learning directly and often generate unpredictable artifacts.

\section{Our Method}

\subsection{Overview}
For the misaligned optical zoom datasets, we denote the low-resolution image as $X \in \mathbb{R}^{H\times W\times 3}$, the high-resolution image as $Y \in \mathbb{R}^{sH\times sW \times 3}$, and the zoom scale as $s$.
During training, our Squared Deformable Alignment Network (SDAN, modified based on SRResNet), denoted as $f_{sdan}(*)$, takes $\{X, Y'\}$ as input, where $Y' \in \mathbb{R}^{H\times W \times 3}$ is the reference image down-sampled from $Y$, and generates zoomed image $\tilde{X} \in \mathbb{R}^{sH\times sW \times 3}$ as output.
During inference, due to the lack of $Y$ and hence $Y'$, $\{X, X\}$ is input into $f_{sdan}(*)$ to generate zoomed results.
The whole pipeline is visualized in Figure~\ref{fig:pipeline}.

\subsection{Squared Deformable Alignment Module}
\begin{figure}[t]
    \centering
    \includegraphics[scale=0.36]{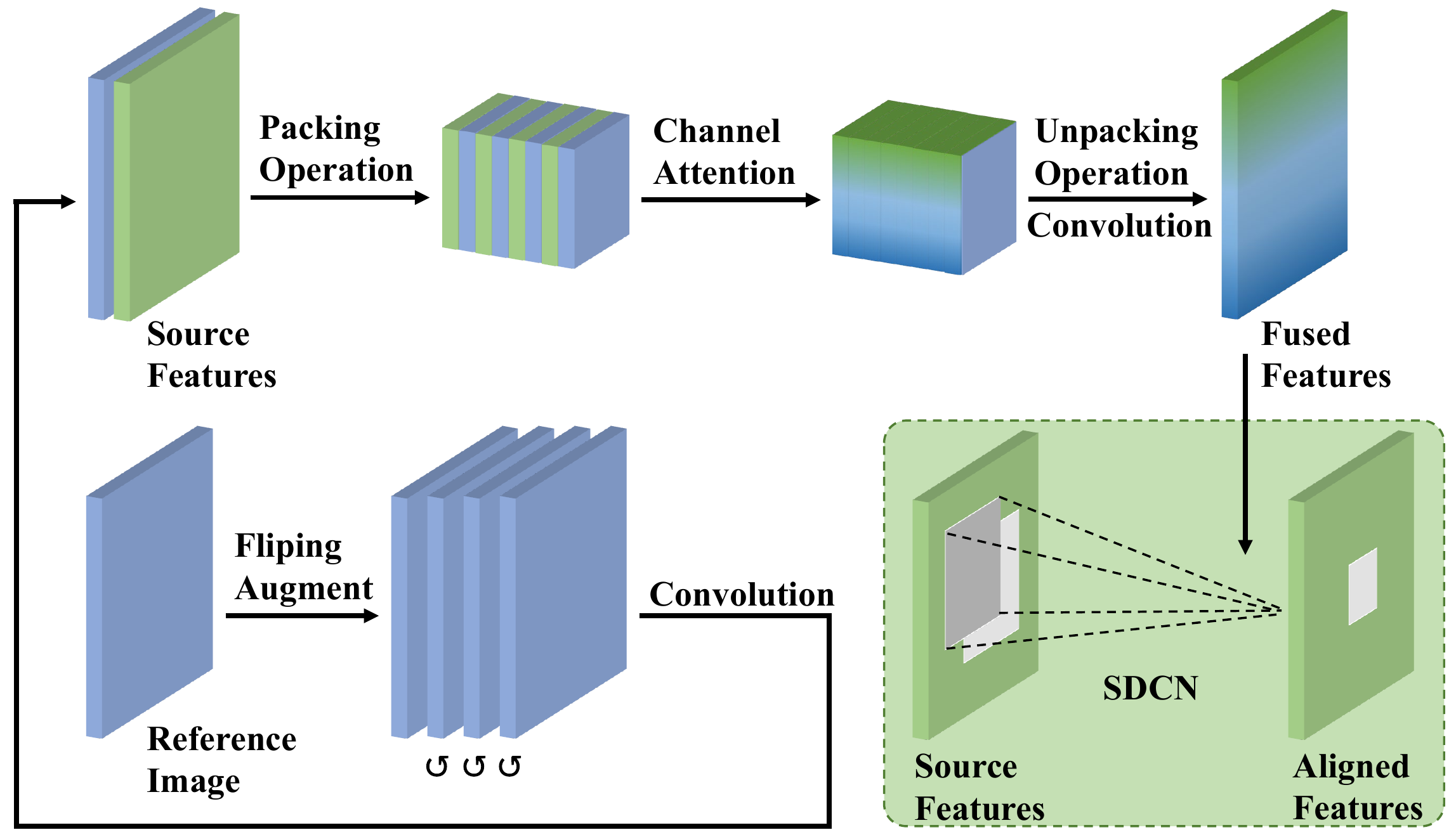}
    \caption{The architecture of squared deformable alignment module for feature alignment. The details of the model can be referred in the Offset Learning Module part.}
    \vspace{-10pt}
    \label{fig:sdcn}
\end{figure}

Squared deformable alignment module is one of the most crucial parts of our SDAN, which learns to align the extracted feature $F^X$ with the extracted reference feature $F^{Y'}$ in an end-to-end manner.
It consists of an offset learning module $f_{offset}(*)$ and a vanilla deformable convolutional network $f_{dcn}(*)$.
Different from the learned offsets in the vanilla DCN~\cite{dai2017deformable} that aim at boosting high-level feature extraction, and the learned offsets in EDVR~\cite{wang2019edvr} that aim at aligning neighboring video frames, the expected offsets in the misaligned optical zoom datasets are usually larger and more difficult to learned.
Another challenge is related to the diversity of the learned offsets, which is claimed to be one of DCN's most important traits that yield better-aligned features, according to Chan et al.~\cite{chan2020understanding}.
However, when DCN is used for inference and the reference image $Y'$ is not present, this trait may lead to overfitting and hence unexpected artifacts in the generated images.
To address the above two issues, we argue that the offset learning module $f_{offset}(*)$ should have as large a receptive field as possible so that it can yield more accurate offsets for feature alignment, and the learned offsets should be smoother with lower diversity.\\

\noindent \textbf{Offset Learning Module.} This module learns to generate offsets from the feature $F^X \in \mathbb{R}^{C\times H \times W}$ and reference feature $F^{Y'} \in \mathbb{R}^{C\times H \times W}$, and its architecture is visualized at the top area of Figure~\ref{fig:sdcn}.
In this module, $F^X$ is extracted from the low-resolution image $X$ via a convolutional layer followed by the ReLUs activation function, and $F^{Y'}$ is extracted from the reference image $Y'$ via three different flip augmentation operations and the same convolutional layer.
Then $F^X, F^{Y'}$ are concatenated together in the channel dimension as $F^{X+Y'} \in \mathbb{R}^{2C \times H \times W}$ for convolution, which aims at extracting spatial connection related features.
To achieve a larger receptive field during convolution, we utilize the packing operation to downsample $F^{X+Y'}$, which folds the spatial feature of $F^{X+Y'}$ into channels. 
To be specific, the packing operation yield features of $(2C \times K \times K) \times \frac{H}{K} \times \frac{W}{K}$, where $K$ is the kernel size of packing.
After the packing operation, we perform channel attention~\cite{hu2018squeeze} to extract channel-related features on the folded features.
Since the packing operation combines the spatial information into channels, the low-cost channel attention can extract both channel-related information and spatial information, and the resulted fused features can better represent the spatial connections.
Later we use the inverse unpacking operation to recover the fused feature back to its original dimension of $2C \times H \times W$.

We name the whole process Cross Packing Attention (CPA), and its motivation comes from our observation that visualized learned offsets from the vanilla DCN tend to be disordered when the misalignment level on images pairs increases. On the other hand, they become more accurate when the number or the kernel size of the offset-learning layers increases.
However, the number or the kernel size of the offset-learning layers cannot be increased infinitely, hence we introduce CPA to increase the receptive field of the offset-learning layers instead, at lower computational costs.
To validate its effectiveness, we have conducted several ablation studies on different settings of the attention layer, and the results shown in Table~\ref{tab:ablation} prove that CPA with flip augmentation achieves the best performance.\\

\noindent \textbf{Feature Alignment Module.} This module aims at aligning the feature $F^X$ by using learned offsets $\Theta$ to generate aligned image features $F^{X'}$.
\begin{equation}
  F^{X'} = f_{dcn}(F^X, \Theta | \mathcal{R}),
\end{equation}
where $f_{dcn}(*)$ is the deformable convolutional netowrk, and $\mathcal{R} =\{(1, 1), (1, 2), \dots (H, W)\}$ is the regular grid sampled from $\Theta$.
After that, the super-resolution module can generate the aligned image $\tilde{X}$ from $F^{X'}$, in which the error gradients caused by misaligned pixels in the pixel-wise loss are greatly reduced.
Specifically, for each point $p_i$, $i \in \{1, 2, \dots, H \times W\}$ in $F^{X'}$ and $F^{X}$, we have:
\begin{equation}
  F^{X'} (p_i) = \sum_{k \in \{1, \dots 9\}} w(p_{i+k}) F^X(p_i + p_{i+k} + \Delta p_{i+k}),
\end{equation}
in which bilinear interpolation is applied to avoid fractional coordinates according to vanilla DCN~\cite{dai2017deformable}.
Note that different from the learned offsets of $(2\times3\times3) \times H \times W$ for each point in the convolutional window of the vanilla DCN, our learned offsets of $2\times H \times W$ are for the whole convolutional window, hence are simpler and more efficient, which we call squared offsets.
What's more, the squared offsets lead to greater quality improvement when $F^{Y'}$ is not available during inference.
The ablation study in Table~\ref{tab:ablation} demonstrates the performance gain over the vanilla DCN.

\subsection{Loss Functions for Misaligned Datasets}
Our ultimate goal is to learn to generate zoomed images $\tilde{X}$ that are similar to the label images $Y$.
However, during our initial experiments, we found that directly optimizing the network $f_{sdan}(*)$ with loss functions $|\tilde{X} - Y|$ led to distorted results.
The reason is that SDCN lacks effective gradients to supervise the offset learning directly.
Therefore we use an additional convolutional layer after SDCN to directly generate temporally aligned results $X' \in \mathbb{R}^{3 \times H \times W}$.
Note that we also take the missing areas due to misalignment into consideration, which is denoted as $M$ and is calculated from the learned offsets on the areas out of the image range.
And $M'$ is the downsampled version of $M$ for aligned $X'$.
The final loss is calculated between $X, X', Y', Y, M, M'$ as:
\begin{equation}
  \mathcal{L} = |X' - Y'| * M' + |\tilde{X} - Y| * M.
\end{equation}
Using the above loss function, our network is optimized to generate the final zoomed images.

\begin{figure*}[htbp]
  \centering
  \includegraphics[scale=0.45]{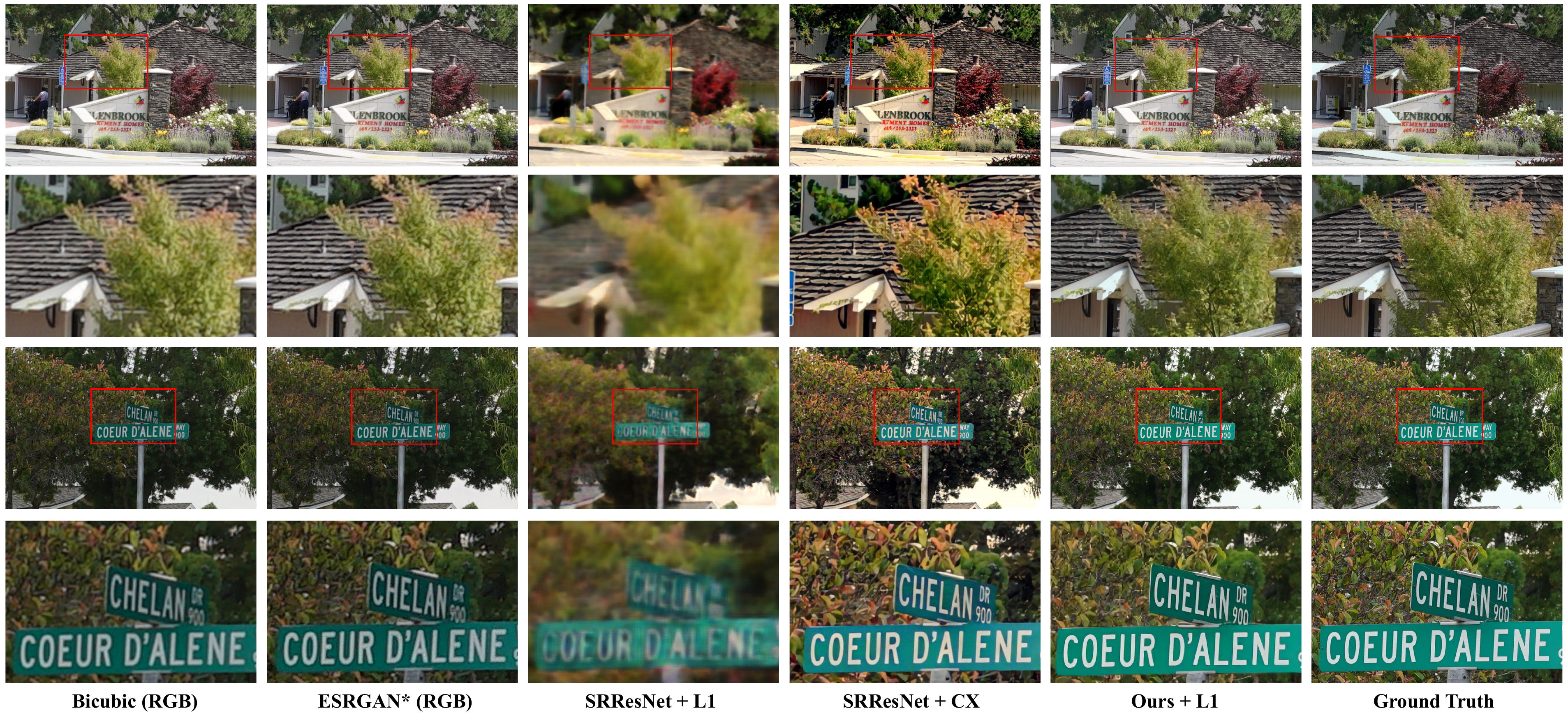}
  \caption{The 4X zoomed results by SOTA methods and ours. \textbf{(Zoom in for best view)}}
  \label{fig:img}
\end{figure*}

\section{Experiments}
\subsection{Dataset}
Our experiments are conducted on the SR-RAW dataset proposed by Zhang et al.~\cite{zhang2019zoom}, which contains 500 image sequences, and each sequence consists of 7 images that captured the same major objects in \{240mm, 150mm, 70mm, 70mm, 50mm, 35mm, 24mm\} focal length settings.
Since the code for 8X scale image loading is not available in the authors' released code, we only trained super-resolution networks on 4X optical zoom, in which each sequence contains 3 low and high resolution pairs for training.
The low-resolution images in RAW format are collected from the camera sensor directly without ISP, while the corresponding high-resolution images in RGB format are processed by the camera ISP.
The misalignment level is varied from 0 pixels to 80 pixels.
During training, we randomly cropped image pairs in $64\times64$ pixels from the low-resolution RAW images ($128\times128$ in RGB) and corresponding magnified patches in $512\times 512$ pixels from the high-resolution images.
The training takes an average time of 5 hours on 8x2080ti GPU cards with 300 epochs.

\subsection{Results}

\noindent \textbf{Quantitative Evaluation.} The performance comparison is showed in Table~\ref{tab:scores}.
Note that the `*' mark after the compared method indicates that it uses a pre-trained model.
For comparison, we utilized the PSNR, SSIM, LPIPS~\cite{zhang2018unreasonable} metrics to measure the differences between the generated zoomed images and the corresponding ground truths.
Because the misalignment commonly exists in the validation set, the evaluation metrics are fair enough to represent the performance of different methods.
Compared with the state-of-the-art method, i.e., SRResNet+CX~\cite{zhang2019zoom}, our method learned on the RAW format achieves the improvement of 3.65 dB higher in PSNR, 0.089 higher in SSIM, and 0.057 lower in LPIPS.
The results also show that learning on the RAW format is superior to learning on the RGB format, and our method learned on the RAW format achieves 1.48 dB higher in PSNR compared with our method learned on the RGB format.
It is worth noting that our method learned on the RGB format also achieves higher performance than [5] on both RAW and RGB formats.
\\

\begin{table}[htbp]
  \vspace{-25pt}
  \caption{Average PSNR/SSIM/LPIPS on 4X optical zoom.}
  \centering
  \begin{tabular}{lcccc}
    \toprule
    &
    \multicolumn{4}{c}{4X Optical Zoom}\\
    \cmidrule(r){2-5}
    & Format &  PSNR \(\uparrow\) & SSIM  \(\uparrow\) & LPIPS \(\downarrow\)\\
    \midrule
    Bicubic & RGB & 20.15 & 0.615 & 0.344 \\
    SRResNet~\cite{ledig2017photo}* & RGB & 23.13 & 0.683 & 0.364 \\
    SRGAN~\cite{ledig2017photo}* & RGB & 20.31 & 0.384 & 0.260 \\
    Perceptual~\cite{johnson2016perceptual} & RAW & 18.83 & 0.354 & 0.270 \\
    ESRGAN~\cite{wang2018esrgan}* & RGB & 22.12 & 0.603 & 0.311 \\
    SRResNet+CX~\cite{zhang2019zoom} & RGB & 22.34 & 0.589 & 0.305 \\
    SRResNet+CX~\cite{zhang2019zoom} & RAW & \textit{26.88} & \textit{0.781} & \textit{0.190} \\
    \midrule
    Ours & RGB & \textbf{29.05} & \textbf{0.858} & \textbf{0.149} \\
    Ours & RAW & \textbf{30.53} & \textbf{0.870} & \textbf{0.133} \\
    \bottomrule
  \end{tabular}
  \label{tab:scores}
\end{table}

\noindent \textbf{Qualitative Evaluation.} Some qualitative results are shown in Figure~\ref{fig:img}.
We use red rectangles to highlight the more obvious areas in the full images and then show the zoomed versions.
From the visualized results, one can easily notice that our method achieves the best visual quality with the most accurate colors, especially in the area contains green plants.
Besides, our method also achieves competitive texture and edge recovering performance compared with the GAN based method and the contextual loss based method.

\subsection{Ablation Study}
In this section, we compared each component of our proposed method with the vanilla DCN to demonstrate our superiority.
The quantitative comparison is shown in Tab~\ref{tab:ablation}.
In the first row, we compared the squared offsets with the per-point offsets, and the result shows our proposed squared offsets significantly increased the quality of the zoomed images.
Then we compare our proposed CPA with the vanilla channel attention in the second row, in which both are based on SDCN.
The result proves the superiority of CPA compared with the channel attention, especially in the LPIPS metric, which achieves 0.086 lower LPIPS.
Finally, we added the flip augmentation to the method based on CPA and SDCN, which is the final version of our proposed SDAN, and we found the result was greatly improved compared to the method without it.

\begin{table}[htbp]
  \vspace{-10pt}
  \caption{Average PSNR/SSIM/LPIPS on ablation studies.}
  \centering
  \begin{tabular}{lccc}
    \toprule
    &
    \multicolumn{3}{c}{4X Optical Zoom}\\
    \cmidrule(r){2-4}
    & PSNR\(\uparrow\) & SSIM  \(\uparrow\) & LPIPS \(\downarrow\) \\
    \midrule
    DCN & 20.91 & 0.752 & 0.281 \\
    SDCN & 26.41 & 0.765 & 0.282 \\
    \midrule
    + CL-Attention & 25.73 & 0.764 & 0.277 \\    
    + CP-Attention & \textit{27.25} & \textit{0.791} & \textit{0.191}   \\
    \midrule
    ++ Flip Augment & \textbf{30.53} & \textbf{0.870} & \textbf{0.133}\\
    \bottomrule
  \end{tabular}
  \label{tab:ablation}
  \vspace{-15pt}
\end{table}

\section{Conclusions}
In this paper, we propose an effective and efficient squared deformable alignment network for misaligned optical zoom learning.
Our proposed squared offsets and cross packing attention enable the network to learn the alignment and optical zoom at the same time to solve the misalignment problem in using pixel-wise loss functions.
Compared with the state-of-the-art methods that utilized the per-point deformable convolution and the contextual loss functions, our method can generate better super-resolution images with fewer visual artifacts.

\bibliographystyle{IEEEbib}
\bibliography{icme2021template}

\end{document}